\documentclass[runningheads]{llncs}
\usepackage{amsmath}
\usepackage{marvosym}
\usepackage{textcomp}  

\usepackage{svg}
\usepackage[T1]{fontenc}
\usepackage{float}
\usepackage{booktabs}
\usepackage{svg}
\usepackage{multirow}
\usepackage{graphicx}
\usepackage{amssymb}
\usepackage{CJKutf8}
\begin{document}
\title{Towards Cross-modal Retrieval in Chinese Cultural Heritage Documents: Dataset and Solution}

\author{%
Junyi Yuan\inst{1}\textsuperscript{*}\orcidID{0009-0001-0160-6878} \and
Jian Zhang\inst{1}\textsuperscript{*}\orcidID{0009-0001-3092-4576} \and
Fangyu Wu\inst{1}\textsuperscript{(\Letter)}\orcidID{0000-0001-9618-8965} \and
Huanda Lu\inst{2}\orcidID{0000-0002-4924-1038} \and
Dongming Lu\inst{3}\orcidID{0000-0002-7228-0887} \and
Qiufeng Wang\inst{1}\textsuperscript{(\Letter)}\orcidID{0000-0002-0918-4606}
}

\renewcommand{\thefootnote}{\fnsymbol{footnote}}
\footnotetext{* These authors contributed equally to this work.}

\authorrunning{J.Yuan  et al.}
%
\institute{Xi'an Jiaotong-Liverpool  University, China \\
\email{\{Junyi.Yuan21,Jian.Zhang22\}@student.xjtlu.edu.cn}\\
\email{\{Fangyu.Wu02, Qiufeng.Wang\}@xjtlu.edu.cn}
\\
\and NingboTech University, China\\
\email{huandalu@nbt.edu.cn}
\\
 \and
Zhejiang University, China\\
\email{ldm@zju.edu.cn}
}

\maketitle     
\markright{Towards Cross-modal Retrieval in Chinese Cultural Heritage Documents: Dataset and Solution}
\begin{abstract}
China has a long and rich history, encompassing a vast cultural heritage that includes diverse multimodal information, such as silk patterns, Dunhuang murals, and their associated historical narratives. Cross-modal retrieval plays a pivotal role in understanding and interpreting Chinese cultural heritage by bridging visual and textual modalities to enable accurate text-to-image and image-to-text retrieval. However, despite the growing interest in multimodal research, there is a lack of specialized datasets dedicated to Chinese cultural heritage, limiting the development and evaluation of cross-modal learning models in this domain. To address this gap, we propose a multimodal dataset named CulTi, which contains 5,726 image-text pairs extracted from two series of professional documents, respectively related to ancient Chinese silk and Dunhuang murals. Compared to existing general-domain multimodal datasets,  CulTi presents a challenge for cross-modal retrieval: the difficulty of local alignment between intricate decorative motifs and specialized textual descriptions. To address this challenge, we propose LACLIP, a training-free local alignment strategy built upon a fine-tuned Chinese-CLIP. LACLIP enhances the alignment of global textual descriptions with local visual regions by computing weighted similarity scores during inference. Experimental results on CulTi demonstrate that LACLIP significantly outperforms existing models in cross-modal retrieval, particularly in handling fine-grained semantic associations within Chinese cultural heritage. The dataset and source code are available at: \url{https://github.com/yyyjjy/CulTi}

\end{abstract}

\section{Introduction}
\begin{figure}[h]
    \centering
    \includegraphics[width=0.85\linewidth]{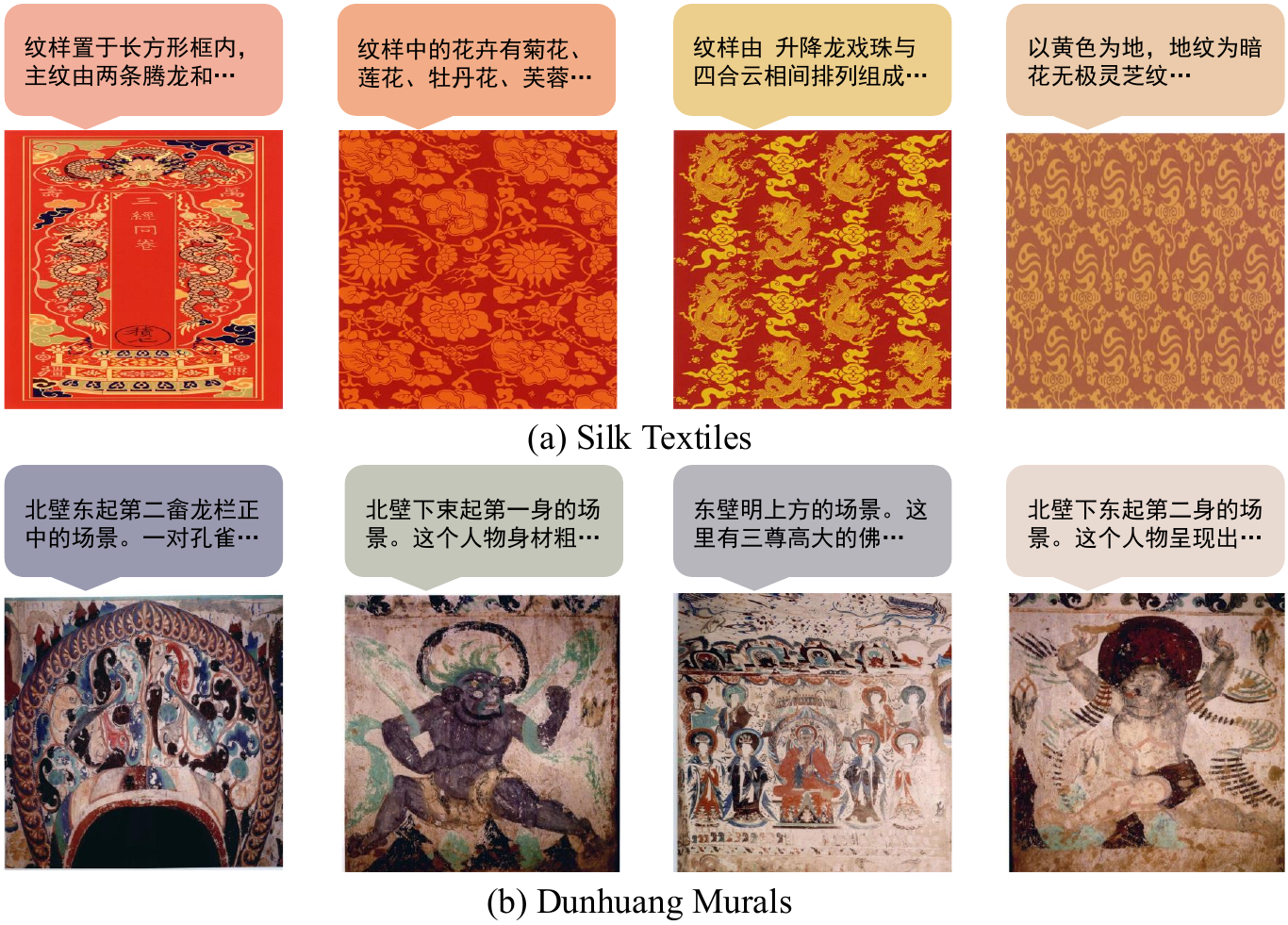}
    \caption{Examples of Chinese Cultural Heritage}
    \label{fig1pdf}
\end{figure}

Cultural heritage, as defined by UNESCO, encompasses both physical artifacts and intangible attributes passed down through generations, ensuring their preservation for the future \cite{nilson2018cultural}. As a country with a long and profound history, China has a rich traditional culture~\cite{liu2018traditional}, where heritage artifacts serve as significant carriers of multimodal information. Among these artifacts, ancient silk~\cite{zuchowska2013china} and the Dunhuang murals~\cite{qu2014conservation} stand out as significant representations of Chinese cultural legacy, as shown in Fig. \ref{fig1pdf}. In practice, textual descriptions of these visual representations are often recorded separately in scriptures or archives. Integrating these pairs of resources provides complementary information, allowing a more comprehensive understanding of cultural heritage. Cross-modal retrieval in Chinese Cultural Heritage Documents aims to establishing bidirectional associations between visual and textual modalities within documents, including archives, artworks, and relic-related records. It encompasses two primary sub-tasks: (1) text-to-image retrieval aims to query the relevant images from documents based on a given textual description, (2) image-to-text retrieval searches for the corresponding textual description from documents based on a given image. By bridging fragmented cultural resources across modalities, this task plays a crucial role in advancing visual question answering, information extraction, intelligent cultural heritage analysis and so on. Despite its significance, research in this area has remained limited, and well-structured datasets are still lacking to support systematic development and evaluation \cite{chen2023rethinking}.

Current Chinese multimodal datasets, such as COCO-CN~\cite{li2019coco} and Flickr30K-CN~\cite{lan2017fluency}, primarily focus on natural scene imagery and lack coverage of Chinese cultural heritage. While several cultural datasets have emerged, including the Dunhuang Grottoes Painting Dataset~\cite{yu2019dunhuang} and the Iranian Cultural Heritage Buildings Dataset~\cite{bahrami2024deep}, they remain unimodal, providing only visual data without aligned textual descriptions. These unimodal datasets fail to support cross-modal retrieval research and limits their utility in contextualizing culturally significant elements. The traditional Chinese art forms, such as silk textiles and murals, contain intricate patterns and layered compositions that hold symbolic significance. The fine-grained local features, like the detailed patterns in textiles or murals, are often overlooked. These limitations significantly constrain their application.

To address this gap, we present the first document-based multimodal dataset CulTi on Chinese cultural heritage, introducing a novel paradigm for cultural data collection. CulTi is derived from publication documents related to ancient Chinese silk and Dunhuang murals, leveraging Optical Character Recognition (OCR)~\cite{nguyen2021survey} tools and ChatGPT-4o~\cite{hurst2024gpt} to obtain image-text pairs. For each image-text pair, the dataset provides four attributes: an ID, title, textual description, and the corresponding image. The final dataset contains 5,726 high-quality image-text pairs in Simplified Chinese, providing valuable resources for multimodal tasks.

\begin{figure*}[h]
    \centering
    \includegraphics[width=0.7\linewidth]{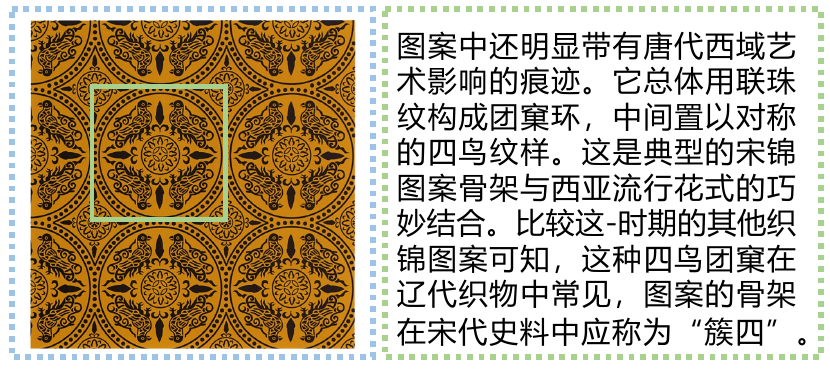}
    \caption{Alignment Between Local Patch and Overall Text}
    \label{exg}
\end{figure*}
While existing multimodal methods~\cite{yang2022chinese,chen2022altclip,gu2022wukong,radford2021learning,xie2023ccmb,pan2023fine} perform well in general cross-modal retrieval tasks, they face significant challenges when dealing with cultural artifacts such as silk patterns and traditional murals. These artifacts typically exhibit rich artistic elements: Dunhuang murals frequently involve complex visual narratives with multi-object compositions and silk patterns often contain repetitive, delicate motifs. As shown in Fig.~\ref{exg}, the image is a repeating pattern, but the text only describes one unit of the pattern, and  even for that single unit, the edge details are ignored, as the text focuses solely on the area within the green frame. Vision-Language Models (VLMs) are deep learning frameworks designed to jointly understand and align visual and textual data~\cite{zhang2024vision}. Existing generic VLMs, such as Chinese-CLIP~\cite{yang2022chinese}, fail to incorporate such information during pre-training and primarily emphasized global alignment, resulting in suboptimal performance in cross-modal retrieval tasks in Chinese Cultural Heritage Documents. To address this limitation, we propose LACLIP, which introduces a Local Visual-Text Alignment module during inference stage of Chinese-CLIP~\cite{yang2022chinese}. By employing weighted alignment between local images and global textual descriptions, LACLIP significantly improves cross-modal retrieval performance, as validated by experimental results.

In summary, our contributions in this work are as follows:
\begin{itemize}
    \item We propose the first multi-modal Chinese cultural heritage dataset for cross-modal retrieval, CulTi, which fills the gap in multimodal data for ancient Chinese culture.
    \item We introduce LACLIP, a fine-tuned Chinese-CLIP model with a localized alignment mechanism. LACLIP improves cross-modal retrieval by incorporating weighted patch similarity to align detailed decorative motifs with textual descriptions.
    \item Extensive experimental results on the CulTi dataset demonstrate that LACLIP enhance  Vision-Language Models (VLMs) in capturing semantic associations between images and texts within the domain of Chinese cultural heritage.
    
\end{itemize}

\section{Related Work}
\subsection{Cross-modal Retrieval Datasets}
Existing datasets for cross-modal retrieval tasks in Chinese primarily focus on general-purpose vision-language tasks. Flickr30K-CN~\cite{lan2017fluency}, containing approximately 31,000 image-caption pairs, is widely adopted for image-text retrieval tasks. COCO-CN~\cite{li2019coco} offers more extensive Chinese annotations, supporting advanced multimodal learning applications. For large-scale model training, the Wukong dataset~\cite{gu2022wukong} provides diverse vision-language pairs, facilitating robust cross-modal alignment. While these datasets have significantly contributed to cross-modal retrieval research, they lack domain-specific knowledge, particularly in the context of cultural heritage preservation.
\\
\indent Additionally, in the realm of cultural heritage, single-modal datasets are developed for specialized tasks. For instance, Chinese Dunhuang Grottoes Painting Dataset~\cite{yu2019dunhuang} focused on tasks such as image restoration and stylistic analysis. Another example is the Iranian Cultural Heritage Buildings Dataset~\cite{bahrami2024deep}, which is designed for defect detection and preservation using advanced deep learning techniques. However, these datasets are limited in addressing challenges like cross-modal retrieval and multimodal fusion, which are crucial for comprehensive cultural heritage preservation. This gap highlights the need for domain-specific multimodal datasets that integrate textual and visual information in a structured manner.
\\
\indent To the best of our knowledge, we introduce the first Chinese multimodal dataset to support the digital preservation and analysis of Chinese cultural heritage, bridging the gap between general-purpose cross-modal datasets and domain-specific single-modal collections. A comparison of datasets is presented in Table~\ref{tab:datasets}.
\begin{table}[h!]
\centering
\caption{Datasets comparison, where "V" is for "Vision", "L" is for "Language", and "–" indicates data missing.}
\label{tab:datasets}
\begin{tabular}{llcccc}
\toprule
Datasets&Modalities&Images&Texts   \\
\midrule
Flickr30K-CN~\cite{lan2017fluency} & V+L & 31,783 &31,783   \\
Wukong~\cite{gu2022wukong} & V+L & 101,483,885 &101,483,885 \\
COCO-CN~\cite{li2019coco} & V+L & 20,342&27,218    \\
Dunhuang Grottoes Painting Dataset~\cite{yu2019dunhuang}&V&600&-\\
Iranian Cultural Heritage Buildings Dataset~\cite{bahrami2024deep}&V&10,500&-\\
CulTi (Ours)&V+L&5,726&5,726\\
\bottomrule
\end{tabular}
\end{table}

\subsection{Cross-modal Retrieval Models}With the release of CLIP~\cite{radford2021learning}, Vision-Language Models (VLMs)~\cite{zhang2024vision} have been widely applied in cross-modal retrieval. Existing Chinese-supported VLMs~\cite{zhang2024vision}, such as AltCLIP~\cite{chen2022altclip}, R2D2~\cite{xie2023ccmb}, and Chinese-CLIP~\cite{yang2022chinese}, are primarily pre-trained on large-scale open-domain datasets, which focus on global alignment between vision and language modalities. Among these, Chinese-CLIP~\cite{yang2022chinese} emerges as a contrastive learning-based multimodal model, specifically designed to align Chinese texts and images, enabling robust understanding and generation of contextually relevant image-text pairs. However, these models lack specialized knowledge in the cultural heritage domain, limiting their effectiveness in tasks requiring fine-grained understanding of domain-specific content. The global alignment approach, while effective for general-purpose tasks, may not sufficiently capture the nuanced relationships between images and text in specialized domains like cultural heritage.

In contrast to global alignment methods, local alignment approaches aim to capture fine-grained correspondences between specific regions of an image and relevant parts of the text. Multimodal alignment methods show success in tasks like visual question answering (VQA)~\cite{antol2015vqa,jahagirdar2021look} and fine-grained image-text retrieval~\cite{wu2024discriminative}. For example, LexVLA~\cite{li2024unified} has been introduced as an interpretable Visual-Language Alignment framework that learns a unified lexical representation with local alignment and an overuse penalty, while SEA~\cite{yin2024sea} has been proposed as a token-level alignment method that leverages vision-language pre-trained models like CLIP~\cite{radford2021learning} to precisely align visual tokens with the embedding space of Large Language Models (LLMs). However, they typically require extensive pretraining, domain-specific annotations, or tailored architectures, limiting their practicality in the cultural heritage domain. 

In existing research, the primary focus is on general cross-modal retrieval tasks, while in this work, we concentrate on improving image-text retrieval performance in the domain of Chinese cultural heritage. By incorporating fine-tuning and the integration of local alignment during inference, we aim to enhance the model's understanding and retrieval capabilities for culturally specific content.
\section{Dataset Construction and Analysis}
\begin{figure*}[t]
\includegraphics[width=\textwidth]{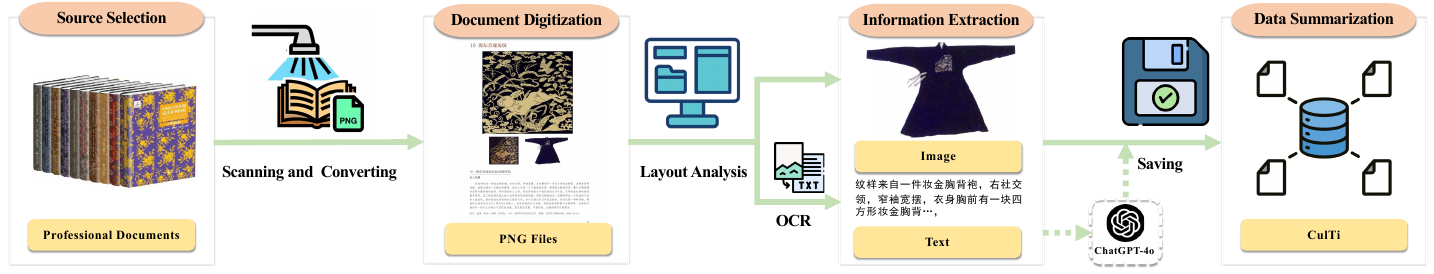}
\caption{General Process of CulTi Construction} \label{CulTi}
\end{figure*}

This section describes the construction of CulTi, involving four main steps: (1) Selecting two authoritative physical publication series as data sources: Original Patterns From Ancient Chinese Textiles Mounting Silks and Complete Collection of Chinese Dunhuang Murals; (2) Digitizing the physical documents using high-precision scanning and converting pages into high-resolution images; (3) Extracting image and text information via cropping and Optical Character Recognition (OCR)~\cite{nguyen2021survey}, with tailored methods for each series to match titles, patterns, and descriptions; (4) Expanding and refining the data, including leveraging  ChatGPT-4o~\cite{hurst2024gpt} to augment brief descriptions and conducting manual proofreading. The general process of dataset construction is shown in Fig.~\ref{CulTi}.
\subsection{Source Selection}
The main sources of data are two series of physical publications called Original Patterns From Ancient Chinese Textiles and  Complete Collection of Chinese Dunhuang Murals. These sources provide a wealth of visual and cultural information, comprehensively demonstrating the essence of traditional Chinese arts and crafts.
\subsubsection{Original Patterns From Ancient Chinese Textiles.}We selected 9 volumes from this series, namely: "Liao and Song Dynasties", "Jin and Yuan Dynasties", "Embroidered Accessories", "Mounting Silks", "Minority Clothing", "Monochrome Woven Silks", "Polychrome Silks", "Velvet and Carpets", and "Painted Images". These volumes cover the Song, Yuan, Ming, and Qing dynasties, each including 150 to 200 patterns. Each pattern in the volumes is accompanied by an image of the original textile, as well as restored and redrawn standard pattern illustrations. The original textile photographs are presented in a smaller size, while the pattern restorations are primarily in vector format. Additionally, relevant textual descriptions detailing its origin, source, and characteristics is included.
\subsubsection{Complete Collection of Chinese Dunhuang Murals.}We selected 11 volumes from this series, namely: "Northern Liang and Northern Wei", "Western Wei", "Northern Zhou", "Sui", "Early Tang", "The Glorious Age of Tang", "Middle Tang", "Late Tang", "Five Dynasties and Song", "Western Xia and Yuan", and the "Mciji Mountain and Bingling Temple" volume. These volumes contain high-definition images of Dunhuang murals, covering the murals from 735 caves in the Mogao Grottoes (spanning 45,000 square meters), 22 caves in the Xiqianfo Cave (818 square meters), and 43 caves in the Yulin Grottoes (5,200 square meters). In addition to the images, vivid explanations by scholars regarding the characteristics of the murals and their historical backgrounds are included. By organically combining  images with interpretations from experts, the volumes provide a comprehensive display of the spiritual world of the Chinese people across different eras.
\subsection{Document Digitization}
In this step, we used a high-precision scanner to transform physical publications into digital files. During the scanning process, we ensured that each page was scanned to maintain clarity for subsequent processing. After scanning, the digital files were typically saved into PDF documents. To facilitate further processing, we converted each page of the PDF documents into PNG image files at a sufficiently high resolution (300 DPI). It allows us to store the content of each page as a separate image file, making it easier to extract both texts and images.
\subsection{Information Extraction}
In this step, we used different techniques: cropping along the white edges and OCR~\cite{nguyen2021survey} technology to extract images and texts, respectively. Due to differences in layout structures and dataset construction requirements between the two series of books, we developed specific information extraction plans for each series.\\
\indent For the Silk series, we extracted titles, original textile images and their descriptions, and pattern images and their descriptions from the pages. We designed different extraction schemes for the varying volumes. For the volumes with numbered titles such as "Liao and Song Dynasties", "Jin and Yuan Dynasties", "Embroidered Accessories", "Minority Clothing", "Monochrome Woven Silks", "Polychrome Silks, and "Velvet and Carpets". We directly used the original page title as the corresponding pattern title in the dataset and combined the original textile name with its historical era to generate the textile  title. Meanwhile, we extracted the relevant parts from the large blocks of descriptive texts for both patterns and textiles, using them as corresponding textual descriptions. Additionally, we added tags like "storage location" and "details" to the physical object descriptions. "Painted Images" has a different structure compared to the previous seven volumes, where the page titles do not have numbers. Thus, we assigned sequence numbers based on the order in which patterns appear in the entire book, attaching these numbers to the beginning of the original title to form the new pattern title. Other processing methods were the same as those used for the first seven volumes. "Mounting Silks" has a unique format where the description of physical objects differs from the other eight volumes, and the pattern and object images are sourced from the mounting sections of ancient paintings and calligraphy. We used the names of the paintings or calligraphy associated with the mounting section as the physical object titles and included related information about the artwork within the physical object descriptions. In total, we extracted 2,495 image-text pairs from the Silk series books. Subsequently, we further cropped and processed the pattern-text pairs extracted from these nine volumes, expanding them into an additional 1,151 pairs.\\
\indent For the Dunhuang Murals series, for each mural in every volume, its corresponding description do not appear on the same page. Thus, we first extracted the textual descriptions from the first half of the volume, converting the original vertical traditional Chinese texts into horizontal simplified Chinese texts with Open Chinese Convert (OpenCC). Then, according to the titles, we extracted the corresponding images from the latter half of the volume. Additionally, locations of murals were added to the original titles to form new titles. After cleaning up empty texts, we obtained 2,080 image-text pairs.

\subsection{Data Summarization}
\begin{figure}[t]
\includegraphics[width=\textwidth]{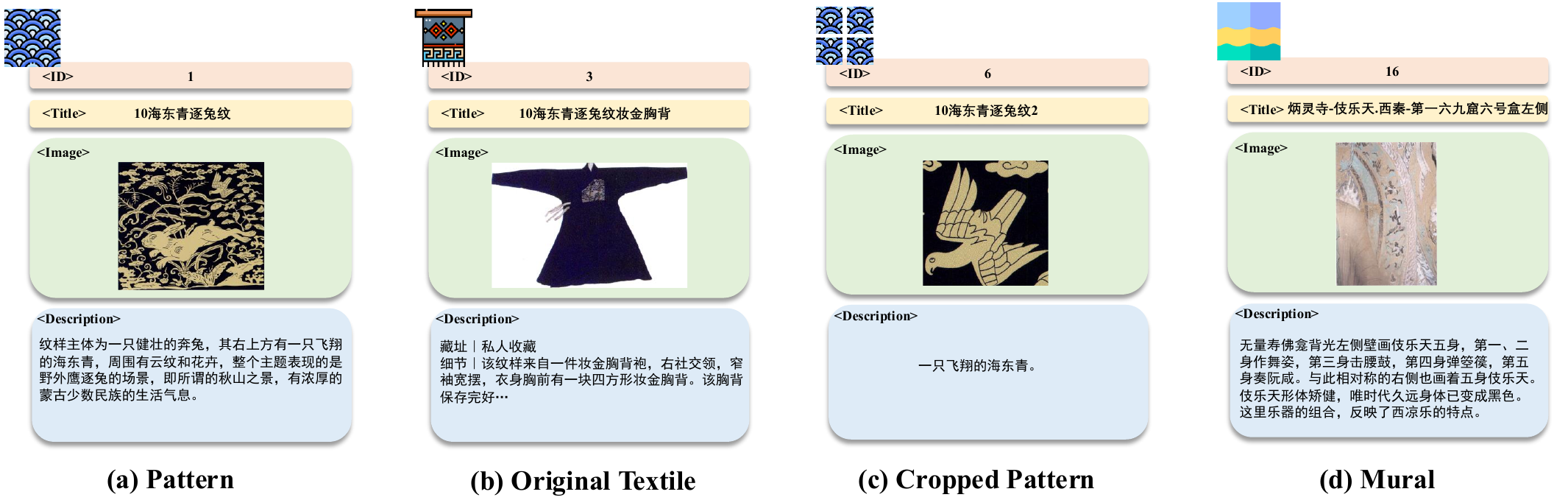}
\caption{Examples of Different Categories of Data in CulTi} 
\label{Examples in CulTi}
\end{figure}
Besides, we used ChatGPT-4o~\cite{hurst2024gpt} to augment 28 pairs of data because this portion of the data lack corresponding textual descriptions for the images. Finally, we structured all data into a same format consisting of ID, title, image, and description, ensures cross-task compatibility for downstream objectives. 

\begin{figure}[t]
\includegraphics[width=1\textwidth]{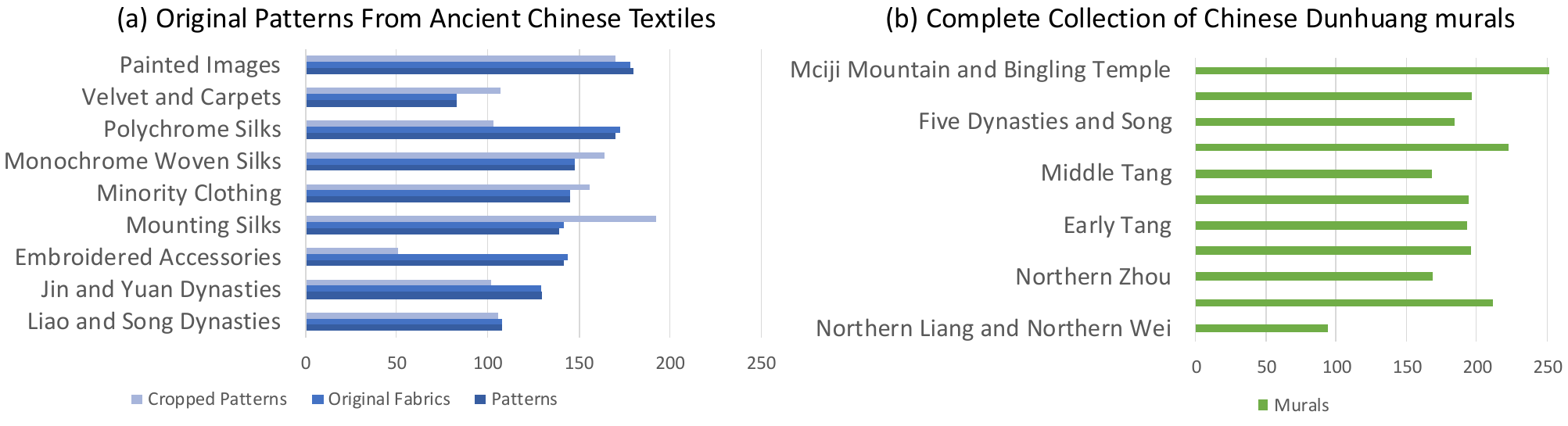}
\caption{Distribution of Different Categories in Each Volume} 
\label{static}
\end{figure}

In summary, we presented CulTi, a high-quality multimodal dataset focused on Chinese cultural heritage, consisting of 5,726 image-text pairs in simplified Chinese. There are four different categories of data as shown in Fig.~\ref{Examples in CulTi}, including pattern, original textile, cropped pattern, and mural. The overall distribution of the dataset from both data sources is illustrated in Fig.~\ref{static}. CulTi includes artifacts from different dynasties and artistic styles, highlighting the diversity of the dataset. Within each volume, data of every category—whether original textiles, patterns, cropped patterns, or murals—is independently partitioned into training, testing, and validation subsets in a 7:1:2 ratio, ensuring a balanced and representative distribution for model development and evaluation.

\section{Method}
\begin{figure}[t]
\includegraphics[width=\textwidth]{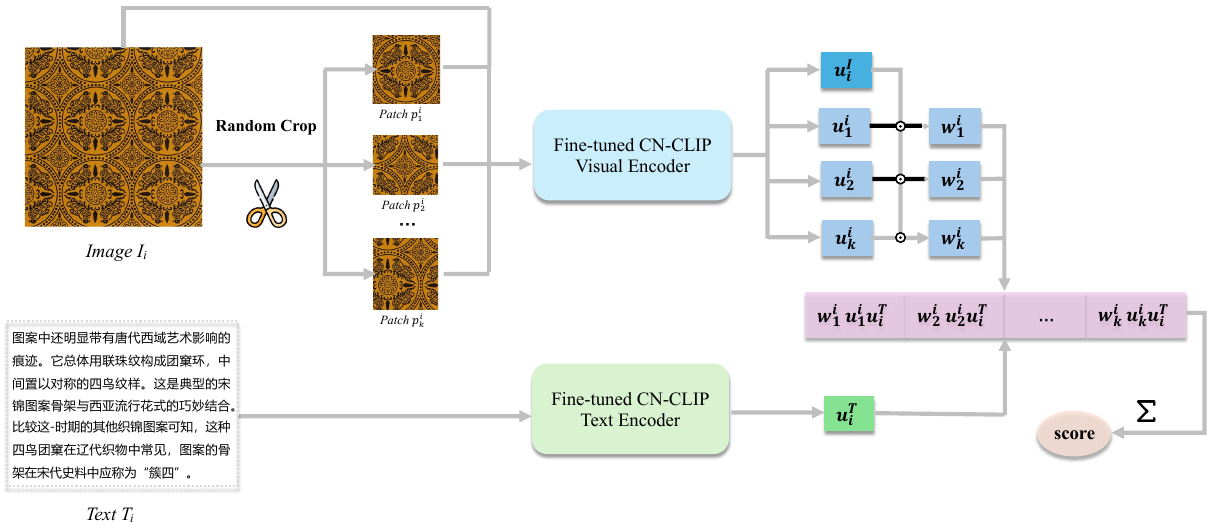}
\caption{Overview of the Proposed LACLIP} \label{model_structure}
\end{figure}

In this section, we present the details of LACLIP model, which aims to enhance the performance of cross-modal retrieval between images and textual descriptions on CulTi dataset. Fig.~\ref{model_structure} shows the detailed structure of LACLIP model, which consists of two key modules, Optimization of Chinese-CLIP and Local Alignment Module.

\subsection{Optimization of Chinese-CLIP}
This module enhances Chinese-CLIP by initializing it with pre-trained weights and fine-tuning it on CulTi for improved image-text alignment. First, we load the pre-trained weights into Chinese-CLIP~\cite{yang2022chinese} to leverage prior knowledge from large-scale multimodal training. The weights are transferred to the text encoder RoBERTa-wwm-Large~\cite{liu2019roberta} and the image encoder ViT-H/14~\cite{dosovitskiy2020image}, providing a robust initialization for the model. Subsequently, the model is fine-tuned on the CulTi training set, thereby enhancing its capability to match images with corresponding textual descriptions.

\subsubsection{Modality Encoding.}
The LACLIP model uses specialized encoders to process the modality information of image-text pairs. For a given image-text pair \(\{T_i, I_i\}\), the textual component \(T_i\) is processed by RoBERTa-wwm-Large~\cite{liu2019roberta} to generate its corresponding feature representation \(f^T_i\), while the visual component \(I_i\) is encoded using ViT-H/14~\cite{dosovitskiy2020image} to produce its respective feature representation \(f^I_i\). Both \(f^T_i\) and \(f^I_i\) belong to the same representation domain \(\mathbb{R}^d\).

\subsubsection{Contrastive Learning for Alignment.}
The feature representations \(f^T_i\) and \( f^I_i \) are then mapped into a shared multimodal space. We implement cosine similarity to measure the similarity between these features, as follows:

\begin{equation}
S(f^T_i, f^I_i) = \frac{f^T_i \cdot f^I_i}{\|f^T_i\| \, \|f^I_i\|},
\label{eq:loss}
\end{equation}
where \(S(f^T_i, f^I_i)\) denotes the cosine similarity between \(f^T_i\) and \( f^I_i\).

To further refine the alignment process, we apply dual formulation which aims to explicitly model the bidirectional relationship between image and text. In cross-modal retrieval task, for a given image, it is essential to identify the most semantically relevant text from a pool of candidates, as follows:

\begin{equation}
L_{i2t} = - \frac{1}{N} \sum_{i=1}^{N} \log \frac{\exp(S(f^I_i, f^T_i) / \tau)}{\sum_{j=1}^{N} \exp(S(f^I_i, f^T_j) / \tau)}, i\neq j,
\label{eq:image_to_text_loss}
\end{equation}
where \( \tau \) is the temperature parameter used to adjust the similarity distribution. A higher \( \tau \) smooths the distribution, while a lower \( \tau \) sharpens it. Conversely, for a given text, to retrieve the corresponding image as follows: 
\begin{equation}
L_{t2i} = - \frac{1}{N} \sum_{i=1}^{N} \log \frac{\exp(S(f^T_i, f^I_i) / \tau)}{\sum_{j=1}^{N} \exp(S(f^T_i, f^I_j) / \tau)}, i\neq j.
\label{eq:text_to_image_loss}
\end{equation}
By incorporating separate loss functions for these two directions, we calculate the total loss as: 
\begin{equation}
L = \frac{1}{2} \left( L_{i2t} + L_{t2i} \right).
\label{eq:total_loss}
\end{equation}


The goal is to minimize the total contrastive loss \( L \), ensuring that the image and text representations are well-aligned in the shared multimodal space for both directions. In this manner, the model optimizes the alignment between image and text modalities by employing distinct loss functions for image-to-text and text-to-image matching, which are subsequently averaged to yield the final loss.

\subsection{Local Alignment Module}
During inference, we introduce a local alignment module to further enhance retrieval accuracy. The design of this module is based on our observation that the textual descriptions in the CulTi dataset typically focus on the main subjects of the images. Therefore, we achieve more accurate alignment between images and textual descriptions by leveraging local-to-global matching. 
\subsubsection{Image Segmentation.}
When extracting image features, the original image will be randomly cropped to multiple local patches \( p^i_k \), where \( k \) denotes the index of the \( k \)-th patch from the \( i \)-th image. The local patch \( p^i_k \) is encoded using the fine-tuned Chinese-CLIP~\cite{yang2022chinese} model to obtain the embedding \( u_k^i\ \). This step ensures that each local patch captures information from diverse regions of the image, including those that may contain the main subject.

\subsubsection{Weight Calculation.}
For each local patch embedding \( u_k^i \), we compute its similarity with the original image embedding \(u^I_i\), which is obtained from the image encoder of the fine-tuned Chinese-CLIP~\cite{yang2022chinese} model, to obtain a weight \( w^i_k \). This weight reflects the degree of association between the local patch \( p^i_k \) and the original image \(I_i\), with patches that are more closely related to the main subject receiving higher weights. Mathematically,
\begin{equation}
w^i_k = \frac{\exp(\alpha \cdot \mathrm{S}(u^i_k, u^I_i))}{\sum_{t=1}^{n} \exp(\alpha \cdot \mathrm{S}(u^i_t, u^I_i))},
\label{eq:softmax_weights}
\end{equation}
where \(\alpha\) is a scaling factor that adjusts the magnitude of the weights. \( \mathrm{S}(u^p_k, u^I_i) \) refers to the similarity between the local patch and original image embeddings.
\subsubsection{Weighted Averaging.}
The final image-text similarity is computed by aggregating the similarity between each local patch and the corresponding text embedding, with the weights reflecting the relevance of each local patch to the main subject of the image. Specifically, we summarize the similarities between all patches and the text embedding to represent the image-text pair, as follows:
\begin{equation}
S_{final} = \sum_{k=1}^{n} w^i_k\cdot \mathrm{S}(u^i_k, u^T_i),
\label{eq:image_text_similarity}
\end{equation}
where \( n \) defines the total number of local patches and \(u^T_i\) is the text embedding obtained from the text encoder of the fine-tuned Chinese-CLIP~\cite{yang2022chinese} model.
\begin{figure*}[t]
\includegraphics[width=\textwidth]{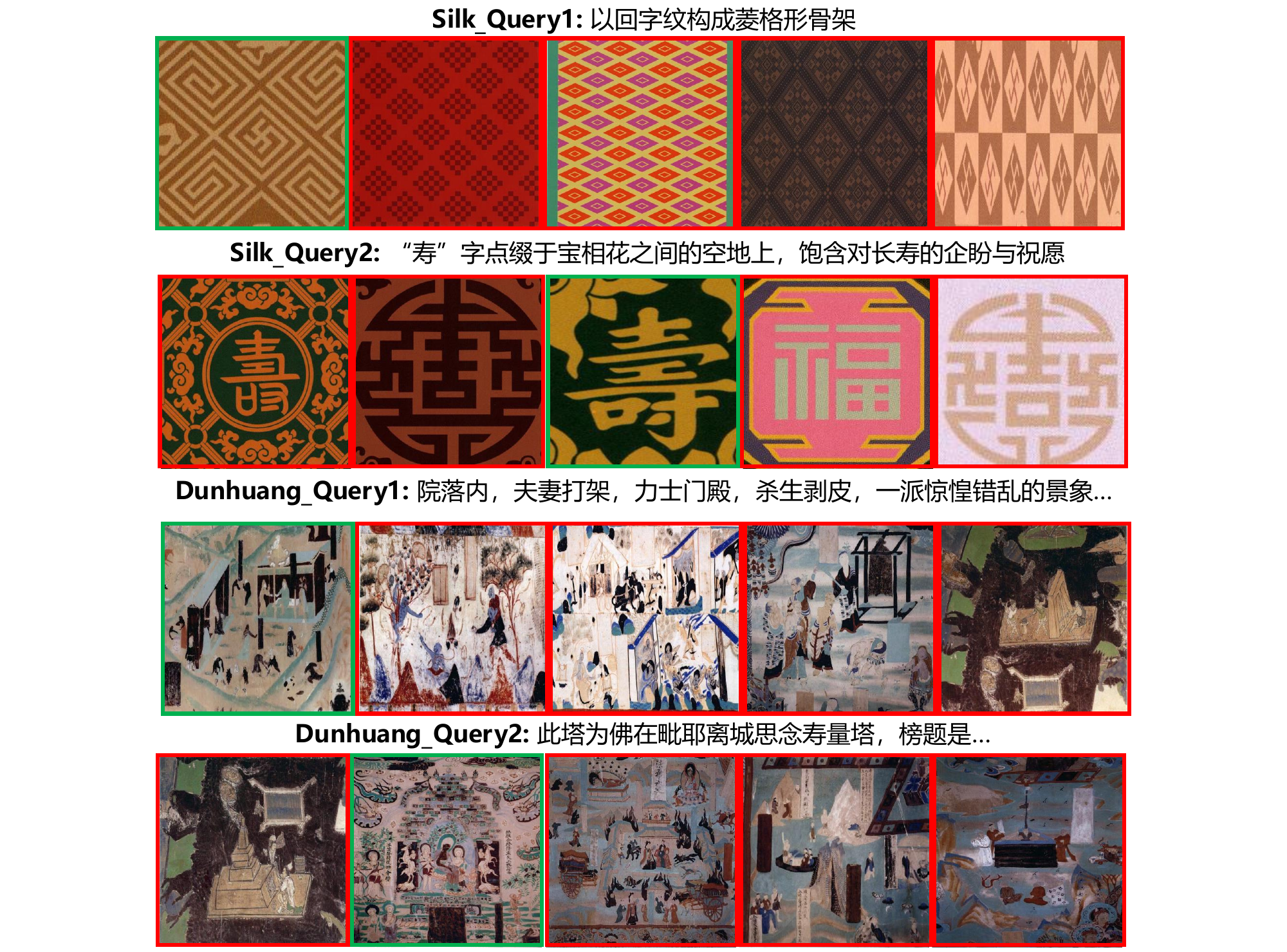}
\caption{Examples of top-5 predictions for text-to-image retrieval. The result in the green frames is correct, while the results in the red frames are incorrect.} \label{t2i}
\end{figure*}

\section{Experiments}
\subsection{Experimental Settings and Baselines}
\subsubsection{Implementation Details.}We employed ViT-H/14~\cite{dosovitskiy2020image} as the visual backbone and utilized RoBERTa-wwm-large~\cite{liu2019roberta} as the textual backbone to construct the Chinese-CLIP~\cite{yang2022chinese} model. The experimental trials were executed utilizing a singular NVIDIA RTX 3090 GPU, endowed with 24 gigabytes of memory. The fine-tuning process was implemented with a batch size of 32, a learning rate set to 5e-5, and trained for 3 epochs. Additionally, the scaling factor \(\alpha\) defined in Eq.~(\ref{eq:softmax_weights}) was set to 1.02, and the learnable parameter \(\tau\) defined in Eq.~(\ref{eq:image_to_text_loss}) and Eq.~(\ref{eq:text_to_image_loss}) was initialized to 1.0.
\subsubsection{Evaluation Metrics.}In the field of information retrieval, performance evaluation is typically conducted using Recall at K (R@K), which quantifies the proportion of queries that are accurately matched within the top K retrieved instances. Higher R@K values signify superior retrieval performance. To comprehensively assess the matching performance of each model on the CulTi dataset, we aggregate the recall values from both image-to-text and text-to-image retrieval tasks, computing the average value as Mean Recall (MR) to provide an overall metric.
\subsubsection{Baselines.}In the experimental setup, two types of settings were employed: zero-shot setting and fine-tuned setting. The zero-shot setting denoted the models without fine-tuning on CulTi, and the fine-tuned setting referred to the models that underwent fine-tuning on CulTi. For zero-shot setting, we compared chosen version of Chinese-CLIP(CN-CLIP)~\cite{yang2022chinese} against different versions of CLIP~\cite{radford2021learning} and Chinese-CLIP~\cite{yang2022chinese}, alongside two additional models: AltCLIP~\cite{chen2022altclip} and R2D2~\cite{xie2023ccmb}. Subsequently, we evaluated the fine-tuned performance of R2D2~\cite{xie2023ccmb} and multiple versions of Chinese-CLIP~\cite{yang2022chinese}. By further incorporating a local alignment module during inference, LACLIP achieved even greater improvement over its baseline version without the module.

\begin{figure*}[t]
\includegraphics[width=\textwidth]{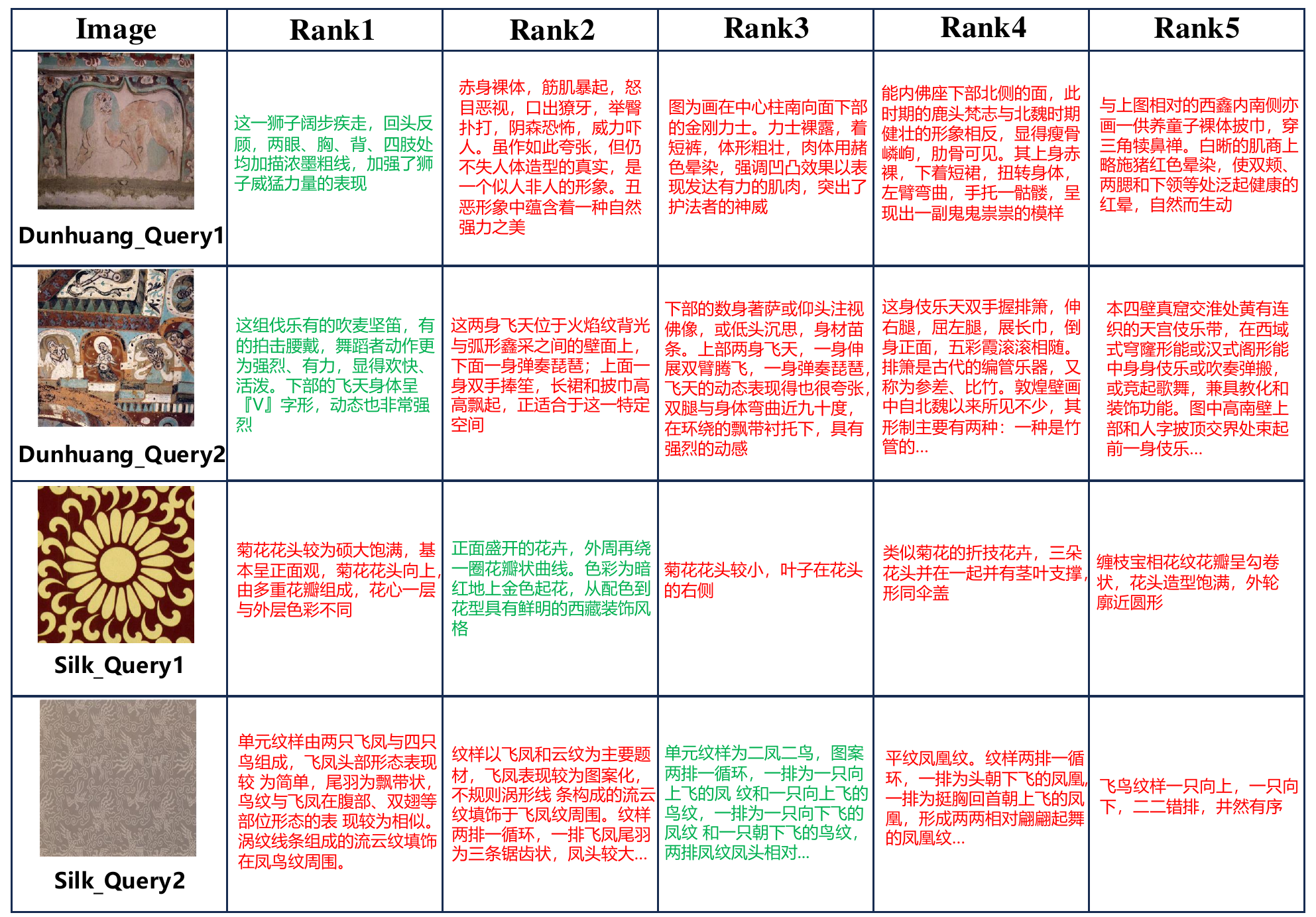}
\caption{Examples of top-5 predictions for image-to-text retrieval. The result in the green frame is correct, while the results in the red frames are incorrect.}
\label{i2t}
\end{figure*}

\subsection{Results and Analysis}

To evaluate the zero-shot retrieval performance of the models, we conducted a comparative analysis. As shown in Table~\ref{tab:performance1}, the $\text{CLIP}_\text{ViT-B}$~\cite{radford2021learning} model achieves an MR of 0.9\% the lowest among all models, indicating limited zero-shot retrieval capabilities. The $\text{CLIP}_\text{ViT-L}$~\cite{radford2021learning} model performs slightly better with an MR of 1.2\%, but still underperforms compared to others. The $\text{AltCLIP}_\text{ViT-L}$~\cite{chen2022altclip} model shows significant improvement with an MR of 14.5\%, demonstrating better zero-shot performance. The $\text{R2D2}_\text{ViT-L}$~\cite{xie2023ccmb} model achieves an MR of 10.8\%, performing moderately but falling short of the top models. The $\text{CN-CLIP}_\text{RN50}$~\cite{yang2022chinese} model, with an MR of 10.0\%, performs comparably to $\text{R2D2}_\text{ViT-L}$~\cite{xie2023ccmb}. The $\text{CN-CLIP}_\text{ViT-B}$~\cite{yang2022chinese} model improves further with an MR of 15.2\%, showcasing stronger zero-shot capabilities. The $\text{CN-CLIP}_\text{ViT-L}$~\cite{yang2022chinese} model achieves an MR of 16.8\%, indicating even better performance. Finally, the $\text{CN-CLIP}_\text{ViT-H}$~\cite{yang2022chinese} model dominates with the highest MR of 22.6\%, establishing itself as the best baseline model for zero-shot retrieval. 
\begin{table}[ht]
\centering
\caption{Zero-Shot Retrieval Performance on CulTi of Different Models}
\label{tab:performance1}
\begin{tabular}{lccccccc}
\toprule
\multirow{2}{*}{Model} & \multicolumn{3}{c}{Text-to-Image} & \multicolumn{3}{c}{Image-to-Text}&MR \\
\cmidrule(lr){2-4} \cmidrule(lr){5-7}
 & R1 & R5 & R10 & R1 & R5 & R10 \\
\midrule
$\text{CLIP}_\text{ViT-B}$~\cite{radford2021learning} & 0.2 & 0.6 & 1.0 &0.0 & 1.1 & 2.2 &0.9\\
$\text{CLIP}_\text{ViT-L}$~\cite{radford2021learning} & 0.0 & 0.9 & 1.6 & 0.7 & 1.5 & 2.4& 1.2\\
$\text{AltCLIP}_\text{ViT-L}$~\cite{chen2022altclip} & 7.0 & 17.1 & 24.5 & 5.2 & 14.1 & 19.1&14.5 \\
$\text{R2D2}_\text{ViT-L}$~\cite{xie2023ccmb} & 3.9 & 12.2 & 18.7 & 2.8 & 10.8 & 16.5&10.8 \\
$\text{CN-CLIP}_\text{RN50}$~\cite{yang2022chinese} & 3.7 & 11.7 & 16.1 & 3.8 & 9.8 & 14.8& 10.0\\
$\text{CN-CLIP}_\text{ViT-B}$~\cite{yang2022chinese} & 5.8 &18.1 & 24.7 & 5.5 & 15.0 & 22.2& 15.2\\
$\text{CN-CLIP}_\text{ViT-L}$~\cite{yang2022chinese} & 7.9 & 20.0 & 27.8 & 5.8 & 16.1&23.3& 16.8\\
$\text{CN-CLIP}_{\text{ViT-H}}$~\cite{yang2022chinese} & \textbf{10.8} & \textbf{28.2} & \textbf{36.9} & \textbf{8.4} & \textbf{20.5} & \textbf{30.5}&\textbf{22.6} \\
\bottomrule
\end{tabular}
\end{table}

As shown in Table~\ref{tab:performance2}, the $\text{R2D2}_\text{ViT-L}$~\cite{xie2023ccmb} model achieves an MR of 39.6\%, showing significant improvement over its zero-shot performance after fine-tuning. The $\text{CN-CLIP}_\text{RN50}$~\cite{yang2022chinese} model, with an MR of 20.8\%, lags behind other fine-tuned models. The $\text{CN-CLIP}_\text{ViT-B}$~\cite{yang2022chinese} model achieves an MR of 38.3\%, demonstrating strong fine-tuned performance. The $\text{CN-CLIP}_\text{ViT-L}$~\cite{yang2022chinese} model improves to an MR of 43.7\%, further solidifying its effectiveness. The $\text{CN-CLIP}_\text{ViT-H}$~\cite{yang2022chinese} model reaches an MR of 47.3\%, nearly matching the top performer and confirming its status as the best baseline model. Finally, our $\text{LACLIP}_\text{ViT-H}$ model achieves the highest MR of 47.9\% as a 0.6\% improvement. This enhancement is directly attributed to the local alignment module incorporated during the inference stage, which better aligns textual features with visual regions, thereby improving retrieval accuracy. Although LACLIP shows a slight decline in the image-to-text retrieval task, its overall performance still surpass even the strong baseline of $\text{CN-CLIP}_\text{ViT-H}$~\cite{yang2022chinese}. 

\begin{table}[t]
\centering
\caption{Fine-tuned Retrieval Performance on CulTi of Different Models}
\label{tab:performance2}
\begin{tabular}{lccccccc}
\toprule
\multirow{2}{*}{Model} & \multicolumn{3}{c}{Text-to-Image} & \multicolumn{3}{c}{Image-to-Text}&MR \\
\cmidrule(lr){2-4} \cmidrule(lr){5-7}
 & R1 & R5 & R10 & R1 & R5 & R10 \\
\midrule

$\text{R2D2}_\text{ViT-L}$~\cite{xie2023ccmb} & 20.0 & 42.2 & 56.0 & 19.7 & 44.2 & 55.3&39.6 \\
$\text{CN-CLIP}_\text{RN50}$~\cite{yang2022chinese} & 7.9 & 24.1 & 32.6 & 7.2 & 21.4 & 31.8& 20.8\\
$\text{CN-CLIP}_\text{ViT-B}$~\cite{yang2022chinese} & 19.7 &43.9 & 54.8 & 17.4 & 40.9 & 53.0&38.3 \\
$\text{CN-CLIP}_\text{ViT-L}$~\cite{yang2022chinese} & 22.9 & 49.6& 62.2 & 21.9 & 46.2&59.3& 43.7\\
$\text{CN-CLIP}_\text{ViT-H}$~\cite{yang2022chinese} &27.2  & 53.2 & 64.5 & \textbf{24.5} & \textbf{50.4} & \textbf{64.0}&47.3 \\
$\text{LACLIP}_\text{ViT-H}$ (Ours)& \textbf{28.1} & \textbf{56.6} & \textbf{66.2} & 
23.6 & 49.9 & 62.9&\textbf{47.9}\\
\bottomrule
\end{tabular}
\end{table}
These results fully demonstrate that our method, by introducing the local alignment module, significantly enhances cross-modal retrieval capabilities, particularly excelling in the text-to-image retrieval task, providing an effective solution for cross-modal retrieval challenges. Specific examples are shown in Fig.~\ref{t2i} and Fig.~\ref{i2t}.

\subsection{Failure Cases}
For the text-to-image retrieval task, as shown in Fig.~\ref{t2i}, although the rank 1 prediction result for Silk\_Query2 is incorrect, it still includes the character \begin{CJK*}{UTF8}{gbsn}"寿"\end{CJK*} (longevity) from the text, which is similar to the correct result. Similarly, for Dunhuang\_Query2, the correct result is returned at rank 2, but the image at rank 1 also includes the \begin{CJK*}{UTF8}{gbsn}"塔"\end{CJK*} (pagoda) mentioned in the query. For the image-to-text retrieval task, in Fig.~\ref{i2t}, Silk\_Query1 returns the correct result at rank 2, but the text description at rank 1 also mentions \begin{CJK*}{UTF8}{gbsn}"菊花"\end{CJK*} (chrysanthemum), which is highly relevant to the pattern in the image. Likewise, for Silk\_Query2, although the results at rank 1 and rank 2 are incorrect, they still describe the phoenix and birds depicted in the image.

These cases indicate that our model, although not always accurate in retrieving the exact match, demonstrates a strong ability to capture and reflect the semantic elements present in the queries.

\section{Conclusion}

In this work, we introduced CulTi, the first multimodal dataset dedicated to Chinese cultural heritage. CulTi comprises 5,726 high-quality image-text pairs extracted from authoritative publications on ancient Chinese textiles and Dunhuang murals. CulTi fulfill the gap in multimodal cultural area provides valuable resources for advancing research in digital cultural heritage preservation and semantic analysis. We also proposed the LACLIP model, which incorporates a local alignment module into a fine-tuned Chinese-CLIP framework. By emphasizing weighted alignment between local visual features and global textual descriptions, LACLIP significantly enhances cross-modal retrieval performance. In future work, we plan to expand the dataset and explore additional tasks, further refining local alignment techniques to boost performance in even more challenging multimodal scenarios.
\begin{credits}
\subsubsection{\ackname}This work was supported by the National Natural Science Foundation of China (No. 62276258), 
the XJTLU Research Development Fund (REF-22-01-002), the Key Technology R\&D Program of Ningbo (2023Z143), 
and the Suzhou Municipal Key Laboratory for Intelligent Virtual Engineering (SZS2022004).
\end{credits}
%
%

\bibliographystyle{splncs04}
\bibliography{main}





\end{document}